# Compliance checking in reified I/O logic via SHACL

Livio Robaldo[1] and Kolawole J. Adebayo[2]

[1]Legal Innovation Lab Wales - Swansea, UK
livio.robaldo@swansea.ac.uk
[2]Dublin City University, Dublin, Ireland
collawolley3@yahoo.com

**Abstract.** Reified Input/Output (I/O) logic[21] has been recently proposed to model real-world norms in terms of the logic in [11]. This is massively grounded on the notion of reification, and it has specifically designed to model meaning of natural language sentences, such as the ones occurring in existing legislation. This paper presents a methodology to carry out compliance checking on reified I/O logic formulae. These are translated in SHACL (Shapes Constraint Language) shapes, a recent W3C recommendation to validate and reason with RDF triplestores. Compliance checking is then enforced by validating RDF graphs describing states of affairs with respect to these SHACL shapes.

## 1  Introduction

Reified I/O logic[21] is a version of standard I/O logic [15] that incorporates the logic in [11]. This is grounded on the notion of *reification*, a well-known technique used in many contemporary approaches to Natural Language Semantics.

Reification is a formal mechanism that associates instantiations of high-order predicates and operators, such as modal or temporal operators, with FOL constants or variables. The latter can be then directly inserted as arguments of other FOL predicates, which may be in turn reified again into new FOL terms. The final resulting formulae are then *flat conjunctions* of atomic FOL predicates.

In reified I/O logic, we build if-then rules in which the antecedent and the consequent are formulae in the logic defined in [11]. There are three sets of these if-then rules: $O$, the set of obligations; $P$, the set of permissions; and $C$, the set of constitutive rules. $\forall (a, b) \in O$ reads as "*given a, b is obligatory*", $\forall (a, b) \in P$ reads as "*given a, b is permitted*", and $\forall (a, b) \in C$ reads as "*given a, b holds*" (in other words, constitutive rules are indeed standard logical implications '$a \rightarrow b$').

If-then rules might be recursively reified again into new FOL terms, allowing to assert meta-properties on the corresponding norms. To the best of the author's knowledge, reified I/O logic is the single logic for normative reasoning proposed so far in the literature allowing so, although past proposals in AI&Law, e.g., [2], [23], has identified this need. These proposals argue that norms must be seen as *objects with properties*, such as jurisdiction (the geographical area in which norms apply), authority (the producers and amenders of the norms, together with their ranking status within the sources of law), temporal properties (the time when the norm has been enacted, the time when it holds, etc.), etc. As shown below in section 2, norms can be even the objects of other norms.

Past research in reified I/O logic has focused on how to *build* formulae associated with norms in natural language. On the other hand, this paper presents an implemented solution to *use* these formulae for compliance checking, i.e., for inferring which norms have been violated in a given state of affairs. Specifically, it proposes a serialization of reified I/O logic formulae in SHACL[1] shapes and rules. These are then executed on RDF graphs describing states of affairs.

As it will be explained below, reification is basically the same mechanism at the basis of both RDF and SHACL. Thus, using these W3C standards to encode and execute reified I/O logic formulae appears to be a straightforward solution.

## 2   Background - Reification and reified I/O logic

Reification is a well-known technique used in linguistics and computer science for representing abstract concepts. These are associated with explicit objects, e.g., FOL terms (see below in this section) or RDF individuals (see section 3 below), on which we can assert (meta-)properties.

Both the framework in [11] and the RDF standard represent knowledge in terms of *flat* lists of (atomic) FOL predicates applied to FOL terms. In RDF, the elements of these flat lists are triples "(subject, predicate, object)", while [11] also allows predicates with higher arity; however, this does not enhance the overall expressivity in that any n-ary predicate can be transformed into an (equivalent) conjunction of binary predicates.

As pointed out above, reified I/O logic formulae are if-then rules in which both the antecedent and the consequent are represented as conjunctions of (reified) predicates. Universal and existential quantifiers are added to bound the free variables occurring in the formulae. Universals that outscope the whole if-then rules are used to "carry" individuals from the antecedent to the consequent.

A simple example from [20] is shown in (1). The if-then rule in (1) encodes in reified I/O logic part of Art.5(1)(a) of the GDPR. The formula belongs to the set *O* (note "∈ *O*" in (1)): it is an obligation requiring each personal data processing to be *lawful*.

---

[1] https://www.w3.org/TR/shacl

(1)    $\forall_{e_p}( \exists_{t_1,z,w,y,x}[ \,(RexistAtTime\ e_p\ t_1) \wedge$
$(PersonalData\ z\ w) \wedge (DataSubject\ w) \wedge$
$(Controller\ y\ z) \wedge (Processor\ x) \wedge (nominates\ y\ x) \wedge$
$(PersonalDataProcessing'\ e_p\ x\ z) \,],$
$(isLawful\ e_p)\,) \in O$

Formulae in reified I/O logic employ two kind of predicates: primed predicates such as *PersonalDataProcessing'* and non-primed predicates such as *DataSubject*. The former are obtained by reifying the latter; the first argument of primed predicates, e.g., $e_p$, is the reification of the non-primed counterpart. Reifications are FOL term explicitly referring to *the fact that* a certain relation holds between certain individuals. For instance, $e_p$ refers to the action of processing carried out by $x$ on $z$, i.e., *the fact that $x$ is processing $z$*.

It is not necessary to reify *all* non-primed predicates occurring in the formulae. In (1), we could have also reified (*DataSubject w*), e.g., transformed it into (*DataSubject' $e_w$ w*), in which $e_w$ would represent "the fact that $w$ is a data subject". However, in order to minimize the variables and so the size of the formulae, predicates are reified only when needed.

Therefore, we do reify (*PersonalDataProcessing x z*) into the predicate (*PersonalDataProcessing' $e_p$ x z*), where $e_p$ explicitly refers to this action of processing, because *we need to assert a property on this action*: in the consequent of the obligation, we require it to be lawful, i.e., to satisfy the *isLawful* predicate. Note that in (1), in order to "carry" the variable $e_p$ from the antecedent to the consequent, a universal quantifier outscoping the if-then rule has been inserted. All other variables are existentially quantified within the antecedent.

The other predicate that $e_p$ is required to satisfy is *RexistAtTime*. This is a special predicates used to assert which reifications "really exist" at a certain time. *RexistAtTime* parallels the well-known predicate *HoldsAt* used in Event Calculus [12]. Its introduction is motivated by the fact that it makes no sense to describe actions and events as true or false. Rather, either they really exist in the real world or they do not. So, for instance, if I want to fly, my wanting action exists while my flying action does not, thus we write[2]:

$$\exists_{e_w,t,e_f}[\,(RexistAtTime\ e_w\ t) \wedge (want'\ e_w\ I\ e_f) \wedge (fly'\ e_f\ I)\,]$$

Thus, formula (1) reads: "for every personal data processing $e_p$ of some personal data $z$, owned by a data subject $w$, controlled by a controller $y$, and processed by a processor $x$ (nominated by $y$), it is obligatory for $e_p$ to be lawful.

---

[2] Further details are available at https://www.isi.edu/ hobbs/bgt-evstruct.text

### 2.1  Using reified I/O logic to model norms from existing legislation

Reified I/O logic models norms in terms of if-then rules between two formulae in the logic defined in [11]. As pointed out in the introduction, also norms need to be reified, in order to assert on them meta-properties such as jurisdiction, authority, temporal properties, etc., or even other norms in a recursive fashion.

In reified I/O logic, norms are reified by introducing *additional* constitutive rules that parallel obligations and permissions and that, respectively, define the corresponding status of being obliged and permitted.

We explain how these additional constitutive rules work with the simple example in (2). [20] discusses more complex examples found in the GDPR, so that the interested reader is addressed to [20].

Sentence (2) exemplifies a norm that occurs as object of another norm: the manager's secretary is obliged to note down *the fact that* the manager is obliged to attend a meeting (if that is the case).

(2)   If a manager is obliged to attend a meeting, his secretary is obliged to note it down in his agenda.

In reified I/O logic, assuming that managers are obliged to attend all meetings that concern their companies, formalized as:

(3)   $\forall_{m_a}\forall_{m_e}(\ ((manager\ m_a) \wedge (meetingAboutCompanyOf\ m_e\ m_a)),$
      $\exists_{e_a,t}[(RexistAtTime\ e_a\ t) \wedge (attend'\ e_a\ m_a\ m_e)]\ ) \in O$

We introduce a parallel constitutive rule stating that, under these conditions, managers are in the state of "being obliged" of attending these meeting. This is stated via a special predicate *ObligedAtTime* that replaces[3] *RexistAtTime*:

(4)   $\forall_{m_a,m_e}(\ ((manager\ m_a) \wedge (meetingAboutCompanyOf\ m_e\ m_a)),$
      $\exists_{e_o,t}[(ObligedAtTime\ e_o\ t\ m_a\ e_a) \wedge (attend'\ e_a\ m_a\ m_e)]\ ) \in C$

(4)    is a constitutive rule (note "∈ C"), i.e., a standard FOL implication. In (4), $e_o$ is the reification of the norm in (3); $e_o$ refers to *the fact that* the manager $m_a$ is obliged to do $e_a$ at time $t$. The obligation $e_o$ can be then used as input for the secretary's obligations. Specifically, sentence (2) is formalized as a new obligation stating that the secretary of every manager in the status of being obliged must note that obligation down in his agenda:

(5)   $\forall_{m_a,t,s,e_o}(\ \exists_e[(manager\ m_a) \wedge (secretaryOf\ s\ m_a) \wedge (ObligedAtTime\ e_o\ t\ m_a\ e)],$
      $\exists_{e_n}[(RexistAtTime\ e_n\ t) \wedge (noteDown'\ e_n\ s\ e_o)]\ ) \in O$

---

[3] *ObligedAtTime* just introduces another modality, alternative to *RexistAtTime*. See https://www.isi.edu/ hobbs/bgt-modality.text for further discussion.

## 2.2   Adding defeasibility to reified I/O logic

The GDPR is grounded on three main principles: lawfulness, fairness and transparency. Specifically, the regulation requires personal data processings to be "processed lawfully, fairly and in a transparent manner in relation to the data subject" (GDPR, Art. 5(1)(a)).

It is well-known that the interpretation of these three principles in contexts is rather difficult, therefore leading to *defeasible* conclusions. As argued in [16], the principle of fairness is still particularly ambiguous and uncertain, and there is not yet a general consensus on how to consistently apply it. For this reason, we will focus below on lawfulness and transparency only, and we will use them to exemplify exceptions and legal interpretations respectively.

(1) showed the obligation referring to the principle of lawfulness. A parallel obligation for transparency is obtained by replacing the predicate *isLawful* with a new predicate *isTransparent*.

**Exceptions in reified I/O logic.** A common pattern in legislation occurs when general rules are overridden by more specific rules in restricted contexts. Those more specific rules are seen as exceptions of the general rule, as penguins may be seen as exceptions of birds with respect to the ability of flying.

In many logics for the legal domain, e.g., [10], exceptions are modelled by introducing special if-then rules called "defeaters" that block general rules in case more specific ones apply. Reified I/O logic implements the very same mechanism. Exceptions are modelled via special predicates "P" that are false by default, i.e., they hold by negation-as-failure (*naf*). *naf*(*P*) is true if either *P* is false or it is *unknown*. On the other hand, when *P* holds, *naf*(*Ex*) is false, and the general rule is blocked. An example, taken from [20], is given by the following rules:

(a) If the data subject has given consent to processing, then this is lawful.
(b) If the age of the data subject is lower than the minimal age for consent of his member state, (a) is not valid.
(c) In the case of (b), if the holder of parental responsibility has given consent to processing, then this is lawful.

(a)-(c) are formalized as the constitutive rules in (6)-(8).

(6)  $\forall_{e_p}( \exists_{t,z,w,y,x}[ (RexistAtTime\ e_p\ t) \wedge (DataSubject\ w) \wedge$
         $(PersonalDataProcessing'\ e_p\ x\ z) \wedge (Controller\ y\ z) \wedge$
         $(Processor\ x) \wedge (nominates\ y\ x) \wedge (PersonalData\ z\ w) \wedge$
         $(GiveConsentTo\ w\ e_p) \wedge naf((exceptionAgeDS\ e_p)) ],$
     $(isLawful\ e_p) ) \in C$

(7)     $\forall_{e_p}( \exists_{t,z,w,y,x,s}[$ (*RexistAtTime* $e_p\ t$) ∧ (*DataSubject w*) ∧
                (*PersonalDataProcessing*' $e_p\ x\ z$) ∧ (*Controller y z*) ∧
                (*Processor x*) ∧ (*nominates y x*) ∧ (*PersonalData z w*) ∧
                (*StateOf s w*) ∧ (< *ageOf*(*w*) *minConsentAgeOf*(*s*)) ],
        (*exceptionAgeDS* $e_p$) ) ∈ C

(8)     $\forall_{e_p}( \exists_{t,z,w,y,x,s,h}[$ (*RexistAtTime* $e_p\ t$) ∧ (*DataSubject w*) ∧
                (*PersonalDataProcessing*' $e_p\ x\ z$) ∧ (*Controller y z*) ∧
                (*Processor x*) ∧ (*nominates y x*) ∧ (*PersonalData z w*) ∧
                (*StateOf s w*) ∧ (< *ageOf*(*w*) *minConsentAgeOf*(*s*)) ∧
                (*hasHolderOfPr h w*) ∧ (*GiveConsentTo h* $e_p$) ],
        (*isLawful* $e_p$) ) ∈ C

**Legal interpretations in reified I/O logic.** Norms can be interpreted in multiple, and often incompatible, ways [14]. Especially when their application scope is wide, as in the case of the GDPR, legislators tend to use *vague* terms, to account for the (wide) range of situations in which the norms must apply.

The GDPR principle of transparency is a rather evident example of vagueness in legal texts, in that it requires controllers to provide information to the data subjects "*in a concise, transparent, intelligible and easily accessible form, using clear and plain language, in particular for any information addressed specifically to a child.*" (GDPR, Art. 12(1)). The words "'clear", 'concise", etc., are of course highly dependent on subjective interpretations.

In case of disputes, judges in courts are in charge of deciding which interpretations are the most appropriate in certain contexts.

Judicial systems usually rank their courts so that the interpretations (and possibly the final verdict) of a court may be overridden by courts of higher rank. For instance, in the UK, the Supreme Court can override the decisions of the Court of Appeal. However, until a final decision is taken by a legal authority that has the power to do it, multiple, and possibly inconsistent, interpretations must be allowed to co-exist within the same knowledge base.

Reasoning is then possible only after certain interpretations are chosen and all those inconsistent with them are discharged. The choice may be enforced by preference criteria, which again rank the interpretations; see, e.g., [22].

In reified I/O logic, legal interpretations are handled by introducing special predicates that parallel those to model exceptions. These predicates refer to the *assumption* that a certain formula, possibly an atomic formula, i.e., a single predicate, is true. While the special predicates to model exceptions are taken by default as false, those to model assumptions are taken by default as true.

(9) shows a simple example. The constitutive rule states that a personal data processing is transparent if every communication about it can be *assumed* to be clear (predicate "*AssumedClear*"). Further predicates can be of course added to the antecedent in (9) for requiring the communication to be (assumed as) concise, appropriate to a child, etc. (9) omits these further assumptions.

(9) $\forall_{e_p}(\forall_{e_c}\exists_{t_1,z,w,y,x}[(RexistAtTime\ e_c\ t_1) \wedge (DataSubject\ w) \wedge$

$(PersonalData\ z\ w) \wedge (Controller\ y\ z) \wedge (Processor\ x) \wedge$

$(nominates\ y\ x) \wedge (PersonalDataProcessing'\ e_p\ x\ z) \wedge$

$(IsAbout\ e_c\ e_p) \wedge (Communicate'\ e_c\ y\ w) \wedge (AssumedClear\ e_c)],$

$(isTransparent\ e_p)\ ) \in C$

The predicate "*AssumedClear*" is true by default. However, someone, e.g., the data subject, can question whether a certain communication was enough clear and ask a judge to decide about that. In case the judge decides that it was not, the controller can appeal to another court, etc.

[20] uses the tag <lrml:Alternatives> of LegalRuleML [1], the XML format used to serialize the reified I/O logic formulae representing GDPR norms, in order to state which legal authorities either *support* or *reject* certain assumptions. Section 5.1 below implements the very same solution in SHACL.

Once this extra-knowledge is encoded, it is possible to select the interpretations out of preference criteria or different rankings of the authorities that support/reject them, as well as defining corresponding inference schema.

The modularity of (reified) I/O logic allows for an easy and scalable enrichment of the formulae with legal interpretations: we can enrich *any* formula "on the fly" by simply conjoining additional predicates that refer to assumptions.

For instance, consider again formula (4) above, stating that every manager is obliged to attend every meeting about his company. Indeed, (4) does not properly reflect real-world scenarios. Very busy managers do not usually attend all meetings about their companies but they tend to skip those with little strategic importance, e.g., those focused on administrative details only.

Thus, a manager should be obliged to only attend meetings in which his presence is relevant, more relevant than in other tasks he should carry out at the same time. This assumption may be spell out by introducing a predicate "*AttendanceAssumedRelevantIn*" as follows:

(10) $\forall_{m_a,m_e}(\ ((manager\ m_a) \wedge (meetingAboutCompanyOf\ m_e\ m_a) \wedge$

$(AttendanceAssumedRelevantIn\ m_a\ m_e)),$

$\exists_{e_o,t}[(ObligedAtTime\ e_o\ t\ m_a\ e_a) \wedge (attend'\ e_a\ m_a\ m_e)]\ ) \in C$

Note that also the relevance of a manager's attendance in a meeting can be highly dependent on subjective interpretations and preferences, for instance by the other meeting's participants. It could be then rather complex for the secretary to "infer" which meetings she should actually note down in the manager's agenda.

## 3   RDF and SHACL for compliance checking

RDF is nowadays *the* W3C standard language for the Semantic Web. Knowledge is represented via sets of triples "(subject, predicate, object)", where predicate is an rdf:Property while subject and object are rdfs:Resource.

Reification is, in essence, the very same mechanism used to represent knowledge in RDF. Each RDF triple itself may be reified into a rdf:Statement[4] referring to the fact that the relation denoted by the predicate holds between the subject and the object. rdf:Statement parallels the solution, exemplified in (3)-(4) above, to reify if-then rules denoting obligations and permissions.

### 3.1   Related works

The first approaches to legal reasoning on RDF triples are dated in the years 2005-2010. Examples are [9] and [4]. These approaches use RDFs/OWL to model the states of affairs and separate knowledge bases of legal rules encoded in special XML formats such as SWRL[5] or LKIF-rules[6]. When executed by suitable legal reasoners, e.g., Carneades (see [8]), these rules perform compliance checking and other legal inferences. In the same spirit, [7] proposed to implement rules in SPARQL[7] while [13] used LegalRuleML [1] to this end.

As pointed out in the previous section, LegalRuleML is also the format used to serialize the if-then rules in [20]. These rules refer to a specific OWL ontology for the data protection domain, i.e., the Privacy Ontology (PrOnto) [17], currently the most exhaustive ontology about the GDPR.

On the contrary, some contemporary approaches, e.g., [3] and [6], propose to encode the legal rules *within* RDFs/OWL itself, and in particular within OWL2 decidable profiles, in order to keep computational complexity under control. Norms are represented as property restrictions, which are special OWL classes. These restrictions refer to the subsets of individuals that comply with the norms.

This paper proposes to encodes legal rules in SHACL, recommended by W3C to validate and make inferences on RDFs/OWL graphs.

SHACL is fully compatible with RDF, in that SHACL instructions are a set of RDF resources (classes and properties). Thus, in SHACL, conditions are provided as constructs expressed in the form of an RDF graph. The present paper formalizes conditional norms, specifically the if-then rules of reified I/O logic, in terms of SHACL instructions/RDF resources.

---

[4] https://www.w3.org/TR/rdf-schema/#ch_*eificationvocab*
[5] https://www.w3.org/Submission/SWRL
[6] http://www.estrellaproject.org/doc/D1.1-LKIF-Specification.pdf
[7] https://www.w3.org/TR/sparql11-overview

SHACL has been scarcely investigated for compliance checking purposes, preliminary works being [18] and [5]. SHACL will be briefly presented in section 5 below. However, before then, the running example used in this paper will be introduced; this is a small OWL ontology drawn from PrOnto.

## 4   A running example

PrOnto is a rather big and detailed ontology. To avoid irrelevant complexity, this paper uses a mini-ontology (with prefix "shRIOL:") inspired by PrOnto. This includes the minimal set of concepts needed for our explanations (see Fig.1).

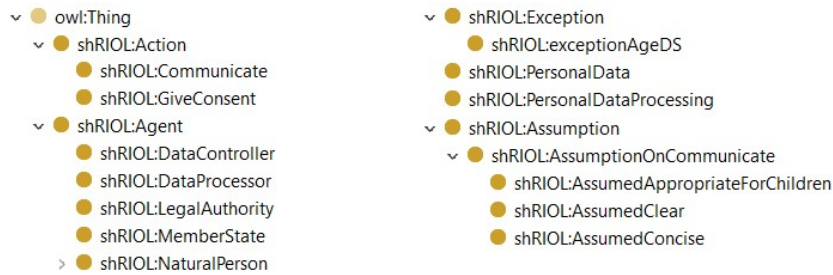

**Fig.1.** The mini-ontology used in this paper.

The main class is PersonalDataProcessing; its instances are events of processing of personal data that must be compliant with the GDPR. On the other hand, the classes Agent and Action respectively specify standard and well-known entities involved in personal data processings (data subject, data controller, etc.) and two actions that may be performed by these agents: GiveConsent and Communicate, the former used to exemplify exceptions, the latter used to exemplify legal interpretations. Finally, the classes Exception and Assumption implement the special predicates used in reified I/O logic to model exceptions and legal interpretations. These classes will be described below in section 5.1.

The ontology includes some restrictions, mirrored from PrOnto. For instance, the class PersonalDataProcessing includes restriction "has-data-controller exactly 1 DataController" stating that each instance is associated through this property with one and only one instance of DataController. This paper omits a detailed description of all restrictions in the ontology in Fig.1.

Section 2.2 above exemplified defeasible reasoning on GDPR principles of lawfulness and transparency in terms of two predicates *isLawful* and *isTransparent*. In PrOnto, as well as in the mini-ontology from Fig.1, these are implemented as homonym *boolean* data properties of PersonalDataProcessing. These will be referred in the SHACL shapes and rules described below in section 5.1.

The model includes three 13-year old data subjects: Hans, Pedro, and Luca, who are respectively from Germany, Spain, and Italy. We assume the three countries have different minimal ages for consent: 16 in Germany, 13 in Spain, and 14 in Italy. Pedro and Luca gave consent to the processing of their personal data; in the case of Hans, the holder of his parental responsibility did. See Fig.2.

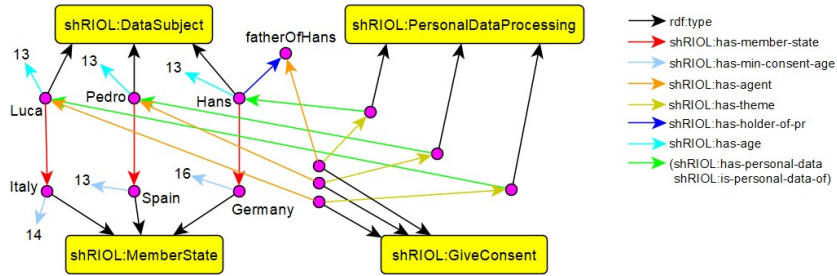

**Fig.2.** Example use case showing consent of processing

On the other hand, Pedro received a communication from his controller. No one ever raised any doubt about that communication, so there is no reason to assume it was not enough clear. Some doubts have been raised instead on a communication that Luca received from his controller. CourtA had to decide about that and it decided that, indeed, it was enough clear. Finally, Hans received two communications from his controller. No one ever raised any doubt about the first one, as in the case of Pedro. Instead, the second communication has been deemed by CourtB to be enough readable and by CourtA to be not. See Fig.3.

In light of GDPR norms, and the corresponding reified I/O logic rules shown above in section 2, and under a (simple) criterion that if at least one legal authority deems at least one communication about some personal data processing not enough readable, then the whole processing is not transparent, we want the enrich the mini-ontology with SHACL shapes and rules able to derive that:

- Hans's and Pedro's processing are lawful while Luca's is not.
- Luca's and Pedro's processing are transparent while Hans's is not.

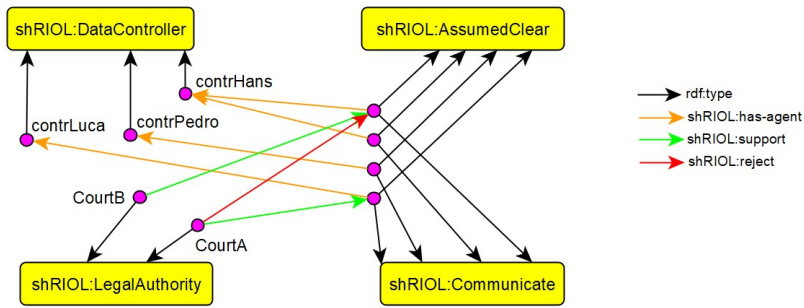

**Fig.3.** Example use case showing clear communication

## 5   SHACL for compliance checking

In 2017, W3C published a new recommendation to validate and reason with RDF triplestores: the Shapes Constraint Language (SHACL)[8]. The use of SHACL in AI is currently a matter of ongoing research (see, e.g., [19]).

SHACL was originally proposed to define special conditions, called "SHACL shapes", against which RDF graphs can be *validated*. However, SHACL "*may be used for a variety of purposes beside validation, including user interface building, code generation and data integration*". In light of this, a current W3C Working Group Note proposes to enrich SHACL shapes with *advanced features*[9] such as
"SHACL rules" to derive inferred triples from asserted ones, prior to validation.

### 5.1   Implementing reified I/O logic in SHACL

This paper proposes to use SHACL for compliance checking with reified I/O logic formulae. In a nutshell, it proposes to:

  - Implement obligations and permissions as SHACL shapes. -
  Implement constitutive rules as SHACL rules.

(11) shows the SHACL shape requiring the boolean attribute is-lawful in the class PersonalDataProcessing to be true. When validated on the ontology in Fig.1, the individuals that do not comply with the SHACL shape in (11) are detected. A parallel shape CheckTransparency is obtained by replacing is-lawful with is-transparent in (11).

(11)          shRIOL:CheckLawfulness rdf:type sh:NodeShape;
                    sh:targetClass shRIOL:PersonalDataProcessing;
                    sh:property [sh:path shRIOL:is-lawful; sh:hasValue "true"];

In (11), "sh:" is the SHACL namespace prefix. (11) is a sh:NodeShape requiring each individual of the sh:targetClass to satisfy the sh:property. The latter constrains the individuals reached from the sh:targetClass through the sh:path to satisfy sh:hasValue.

**Reasoning with Exceptions.** The exception formalized above in the reified
I/O logic formulae in (6), (7), and (8) is implemented via the SHACL rules in (12), (13), (14), and, below, (15). The sh:targetClass of all these SHACL rules is shRIOL:GiveConsent, subclass of shRIOL:Action.

---

[8] https://www.w3.org/TR/shacl
[9] See https://www.w3.org/TR/shacl-af

(12) sh:rule [rdf:type sh:TripleRule; sh:order 0; sh:subject sh:this; sh:predicate
shRIOL:has-min-consent-age; sh:object [sh:path (shRIOL:has-theme shRIOL:has-personal-data shRIOL:is-personal-data-of shRIOL:has-member-state shRIOL:has-min-consent-age)] ]

(13) sh:rule [rdf:type sh:TripleRule; sh:order 1; sh:condition [
         sh:property[sh:path shRIOL:has-min-consent-age; sh:minCount 1]; sh:property
         [sh:path (shRIOL:has-agent shRIOL:has-age);
                    sh:lessThan shRIOL:has-min-consent-age]];
     sh:subject [sh:path shRIOL:has-theme];
     sh:predicate rdf:type; sh:object
     shRIOL:exceptionAgeDS ]

(14) sh:rule [rdf:type sh:TripleRule; sh:order 2; sh:condition [ sh:not [sh:property
     [sh:path shRIOL:has-theme; sh:class shRIOL:exceptionAgeDS]]];
     sh:subject [sh:path shRIOL:has-theme;];
     sh:predicate shRIOL:is-lawful;
     sh:object "true"^^xsd:boolean; ];

Rules are executed according to the sh:order, from the lowest to the highest value. Each rule creates a new triple in the ontology: the sh:predicate is asserted between the sh:subject and the sh:object. These may be the sh:targetClass itself (keyword "sh:this") or a resource reachable from the sh:targetClass through sh:path.

Thus, (12) is executed as first. This rule asserts has-min-consent-age on GiveConsent's instances, while setting its value to the minimal consent age of the Member State of the DataSubject owning the PersonalData processed in the PersonalDataProcessing that is the theme of the GiveConsent instances (see sh:path on sh:object in (12)).

Then, rule (13) compares the minimal consent ages just asserted by (12) with the agents' age. When the latter is lower than (sh:lessThan) the former, the instance of PersonalDataProcessing in the has-theme property of the instance of GiveConsent is asserted as individual of the class exceptionAgeDS.

Then, rule (14) sets as true the property is-lawful of the instances of PersonalDataProcessing that do not (sh:not) also belong to exceptionAgeDS. Finally, (15) implements the reified I/O logic formula (8) above:

(15) sh:rule [rdf:type sh:TripleRule; sh:order 2; sh:condition [sh:property [sh:lessThan
     shRIOL:has-min-consent-age; sh:path (shRIOL:has-theme; shRIOL:has-personal-data
     shRIOL:is-personal-data-of shRIOL:has-age)]];
         sh:property [sh:path (shRIOL:has-theme shRIOL:has-personal-data shRIOL:is-
                 personal-data-of shRIOL:has-holder-of-pr); sh:equals shRIOL:has-
                 agent]];
     sh:subject [sh:path shRIOL:has-theme;];
     sh:predicate shRIOL:is-lawful; sh:object
     "true"^^xsd:boolean; ];

If the age of the DataSubject is lower than the minimal consent age of his/her Member State *and* the agent of the GiveConsent instance is the holder of the DataSubject's parental responsibility, then is-lawful is again set to true.

When executed on the model in Fig.2, the SHACL shape and rules in (11), (12), (13), (14), and (15) properly infer that Hans's and Pedro's personal data processings are lawful while Luca's is not.

**Reasoning with Legal Interpretations.** Formula (9) above encodes the definition of transparent personal data processing adopted in this paper: if all communications about it are assumed to be clear, the processing is transparent.

The easiest implementation of this definition in SHACL is to set the boolean is-transparent of each instance of PersonalDataProcessing as true *by default*, unless another rule (with higher priority) sets it as false. This rule will search if there is at least one Communication about this processing that has been rejected by some LegalAuthority. The rules are shown in (16) and (17).

(16) sh:rule [ rdf:type sh:TripleRule; sh:order 1; sh:condition[sh:property[sh:path shRIOL:is-
     transparent; sh:maxCount 0]];
         sh:subject sh:this;
         sh:predicate shRIOL:is-transparent;
         sh:object "true"^^xsd:boolean; ];

(17) sh:rule [ rdf:type sh:TripleRule; sh:order 0; sh:condition[sh:property[sh:path(shRIOL:is-
     theme-of shRIOL:is-rejected-by) sh:minCount 1]];
         sh:subject sh:this;
         sh:predicate shRIOL:is-transparent;
         sh:object "false"^^xsd:boolean;];

(17) is executed as first. To check if there is at least one LegalAuthority that rejects the assumption on the processing's clearness (note that all instances of PersonalDataProcessing are also instances of AssumedClear, see Fig.3), (17) uses the SHACL instruction "sh:minCount 1" on the is-rejected-by property.

If, after the application of (17), the value of is-transparent is still unknown (i.e., sh:maxCount 0; is true), (16) sets the boolean as true.

The criterion adopted above to determine transparency of a personal data processing is of course a simple one, which does not indeed reflect real-world scenarios. For instance, if CourtB is the UK Supreme Court and CourtA is the UK Court of Appeal, both Hans's personal data processings would be transparent, as the interpretation of CourtB would override the one of CourtA.

These advanced criteria may be easily implemented by introducing further properties and corresponding SHACL rules. On the other hand, it is easily possible to list all legal authorities that reject certain legal interpretations, so that the user is provided with an *explanation* of why his processing is non-transparent.

## 6    Conclusions

Reified I/O logic is a recent deontic framework explicitly designed to represent norms occurring in existing legislation such as the GDPR. This paper investigates reasoning in reified I/O logic, specifically it proposes to model regulative rules as SHACL shapes and constitutive rules as SHACL rules.

The solution proposed here is alternative to some recent approaches to model compliance checking in RDFs/OWL, e.g., [6] and [3]. However, these approaches are silent about incorporating exceptions and legal interpretations in their inference schema, which is instead a central topic in this paper.

Defeasibility is handled by using SHACL operators for negation-as-failure and for establishing an execution order among the rules (sh:order). Mirroring these inferences in native RDFs/OWL seems to be more difficult in that RDFs/OWL does not include corresponding non-monotonic operators.

All OWL, SHACL, and Java files modelling and executing the examples shown in this paper are available on GitHub[10].

## References

1. Athan, T., Governatori, G., Palmirani, M., Paschke, A., Wyner, A.: LegalRuleML: Design Principles and Foundations. Springer International Publishing (2015)
2. Balke, T., da Costa Pereira, C., Dignum, F., Lorini, E., Rotolo, A., Vasconcelos,W., Villata, S.: Norms in MAS: Definitions and Related Concepts. In: Normative Multi-Agent Systems, Dagstuhl Follow-Ups, vol. 4, pp. 1–31 (2013)
3. Bonatti, P.A., Ioffredo, L., Petrova, I.M., Sauro, L., Siahaan, I.S.R.: Real-timereasoning in OWL2 for GDPR compliance. Artificial Intelligence **289** (2020)
4. Ceci, M.: Representing judicial argumentation in the semantic web. In: Casanovas,P., Pagallo, U., Palmirani, M., Sartor, G. (eds.) AI Approaches to the Complexity of Legal Systems (AICOL 2013). Lecture Notes in Computer Science, vol. 8929
5. Debruyne, C., Pandit, H.J., Lewis, D., O'Sullivan, D.: Towards generating policycompliant datasets. In: 13th IEEE International Conference on Semantic Computing (ICSC 2019). pp. 199–203. IEEE (2019)

---

[10] https://github.com/liviorobaldo/shRIOL

6. Francesconi, E., Governatori, G.: Legal compliance in a linked open data framework. In: Araszkiewicz, M., Rodr´ıguez-Doncel, V. (eds.) Legal Knowledge and Information Systems (JURIX 2019)
7. Gandon, F., Governatori, G., Villata, S.: Normative requirements as linked data. In: Wyner, A.Z., Casini, G. (eds.) Legal Knowledge and Information Systems.
8. Gordon, T.: Combining rules and ontologies with Carneades. In: Bragaglia, S.,Dama´sio, C.V., Montali, M., Preece, A.D., Petrie, C.J., Proctor, M., Straccia, U. (eds.) Proceedings of the 5th International RuleML2011@BRF Challenge
9. Gordon, T.F.: Constructing legal arguments with rules in the legal knowledgeinterchange format (LKIF). In: Casanovas, P., Sartor, G., Casellas, N., Rubino, R. (eds.) Computable Models of the Law. Springer (2008)
10. Governatori, G., Olivieri, F., Rotolo, A., Scannapieco, S.: Computing strong andweak permissions in defeasible logic. Journal of Philosophical Logic **6**(42)
11. Hobbs, J., Gordon, A.: A formal theory of commonsense psychology, how peoplethink people think. Cambridge University Press (2017)
12. Kowalski, R., Sergot, M.: A logic-based calculus of events. New Generation Computing **4**(1), 67–95 (1986)
13. Lam, H., Hashmi, M., Scofield, B.: Enabling reasoning with legalruleml. In: Alferes,J.J., Bertossi, L.E., Governatori, G., Fodor, P., Roman, D. (eds.) Rule Technologies. Research, Tools, and Applications - 10th International Symposium, RuleML 2016, Stony Brook, NY, USA, July 6-9, 2016. Proceedings. Lecture Notes in Computer Science, vol. 9718, pp. 241–257. Springer (2016)
14. MacCormick, N., Summers, R.: Interpreting Statutes: A Comparative Study. Applied legal philosophy, Dartmouth (1991)
15. Makinson, D., van der Torre, L.W.N.: Input/output logics. Journal of PhilosophicalLogic **29**(4), 383–408 (2000)
16. Malgieri, G.: The concept of fairness in the gdpr: A linguistic and contextual interpretation. In: Proc. of the 2020 Conference on Fairness, Accountability, and Transparency. p. 154–166. ACM (2020)
17. Palmirani, M., Martoni, M., Rossi, A., Bartolini, C., Robaldo, L.: Pronto: Privacyontology for legal compliance. In: Proceedings of the 18$^{th}$ European Conference on Digital Government (ECDG) (October 2018)
18. Pandit, H.J., O'Sullivan, D., Lewis, D.: Exploring GDPR compliance over provenance graphs using SHACL. In: Khalili, A., Koutraki, M. (eds.) Proc. of the Conference on Semantic Systems (SEMANTiCS 2018).
19. Pareti, P., Konstantinidis, G., Mogavero, F., Norman, T.J.: SHACL satisfiabilityand containment. In: Pan, J.Z., et al. (eds.) Proc. of 19th International Semantic Web Conference (ISWC)
20. Robaldo, L., Bartolini, C., Palmirani, M., Rossi, A., Martoni, M., Lenzini, G.: Formalizing GDPR provisions in reified I/O logic: the DAPRECO knowledge base. The Journal of Logic, Language, and Information **29** (2020)
21. Robaldo, L., Sun, X.: Reified input/output logic: Combining input/output logicand reification to represent norms coming from existing legislation. The Journal of Logic and Computation **7** (2017)
22. Rotolo, A., Governatori, G., Sartor, G.: Deontic defeasible reasoning in legal interpretation: Two options for modelling interpretive arguments. In: Proc. of the 15th International Conference on Artificial Intelligence and Law (ICAIL 2015).
23. Rotolo, A., Tamargo, L., Martinez, D., Villata, S., van der Torre, L., Calardo, E., Governatori, G., Robaldo, L., Palmirani, M.: MIREL D1.2.